\documentclass[
]{ceurart}
\usepackage{listings}
\usepackage[table]{xcolor}
\usepackage{array}

\definecolor{addcol}{HTML}{EAD2D8}      % dusty pink
\definecolor{remcol}{HTML}{DDE5EA}      % blue-gray
\definecolor{subcol}{HTML}{E7DED6}      % taupe
\definecolor{selcol}{HTML}{DDE9DF}      % sage
\definecolor{basecol}{HTML}{EFEFEF}     % light gray

\newcommand{\opadd}{\colorbox{addcol}{\textsf{\scriptsize ADD}}}
\newcommand{\oprem}{\colorbox{remcol}{\textsf{\scriptsize REMOVE}}}
\newcommand{\opsub}{\colorbox{subcol}{\textsf{\scriptsize SUBSTITUTE}}}
\newcommand{\opsel}{\colorbox{selcol}{\textsf{\scriptsize SELECT}}}
\newcommand{\opbase}{\colorbox{basecol}{\textsf{\scriptsize BASE}}}

\begin{document}

\copyrightyear{2026}
\copyrightclause{Copyright for this paper by its authors.
  Use permitted under Creative Commons License Attribution 4.0
  International (CC BY 4.0).}

\conference{CLEF 2026: Conference and Labs of the Evaluation Forum, September 21--24, 2026, Jena, Germany}

\title{Multimodal Sexism Identification and Characterization using Large Language Models and Gradient Boosting }

\title[mode=sub]{Notebook for the EXIST Lab at CLEF 2026}

\author[1]{Kyriakos Chaviaras}[orcid=0009-0006-0963-0377,email=el21157@mail.ntua.gr]\address[1]{Artificial Intelligence and Learning Systems Laboratory \\ School of Electrical and Computer Engineering \\ National Technical University of Athens}
\author[1]{Maria Lymperaiou}[orcid=0000-0001-9442-4186,email=marialymp@ails.ece.ntua.gr]
\author[1]{Athanasios Voulodimos}[orcid=0000-0002-0632-9769,email=thanosv@mail.ntua.gr]

%\cortext[1]{Corresponding author.}
%\fntext[1]{These authors contributed equally.}


\maketitle
\begin{abstract}
We present the AILS-NTUA submission to the EXIST 2026 Lab at CLEF, addressing multimodal sexism identification and characterization in memes (Task 2) and short-form videos (Task 3). Our system follows a feature-engineered late-fusion pipeline built around gradient-boosted regression models and hierarchical post-processing. For memes, we combine visual, textual, demographic, biometric, and LLM-derived semantic indicators designed to capture high-level cues such as stereotyping, objectification, irony, and misogyny. For videos, we investigate the effect of feature selection, frame-based visual representations, OCR-based textual features, acoustic descriptors, and sensor-derived metadata. Development results show that focused LLM-derived semantic cues improve meme sexism identification, while video performance is highly sensitive to feature dimensionality and cross-modal noise. %Official test results further reveal a dev--test discrepancy for videos, where the unfiltered representation generalized better than aggressively selected feature subsets. 
%For videos, development results favor compact feature selection, but official test results show that this conclusion does not fully transfer to unseen data, where the unfiltered representation generalizes better. However, as the official test results show, this conclusion does not fully transfer to unseen data, where the unfiltered representation generalizes better. However, as discussed in the official test results, this conclusion does not fully transfer to the unseen test distribution.
For videos, development results favor compact feature selection, but official test results show that this conclusion does not fully transfer to unseen data, where the unfiltered representation generalizes better.
Overall, our findings highlight the usefulness of targeted semantic feature engineering for static memes and the need for more robust temporal modeling in noisy short-form video settings.
\end{abstract}

\section{Introduction}
Sexism and gender discrimination remain among the most persistent social challenges, manifested over time through the establishment of stereotypes, objectification, reinforcement of inequality and most recently, online harassment \cite{glick1996ambivalent, fredrickson1997objectification, misogyny, ijerph20105770}. The widespread integration of social media into everyday life has created easily accessible channels for gender-related  harassment and discrimination, often indirectly instantiated using humor, irony or sarcasm within multimodal content \cite{ford2008more, drakett2018old}. Typical instances of such are found in memes and short videos, offering  rapid and low-effort mechanisms for opinion dissemination, tailored to the rapid information consumption that characterizes contemporary social media use \cite{memes-digital, milner-meme}. The scale of the phenomenon is non-negligible: studies involving women journalists reveal that 73\% of respondents reported online violence, including direct mental health consequences, while 41\% of respondents reported being targeted by online attacks related to disinformation \cite{posetti2021chilling}.

The multimodal nature of such online content has motivated the integration of related AI techniques that jointly consider text, image, video and audio to offer detection signals in the presence of gender-related discriminatory and offensive cues. 
%The first step towards such endeavors requires the introduction of appropriate benchmarks that challenge the current state of available multimodal architectures. 
Probing cross-modal interactions is central in these tasks, since separate modalities may offer signals of varying intensity; thus, in cases where only a single modality is considered, the overall presence of discriminatory characteristics may be obscured.
%and can be harnessed for post-training to enhance their baseline performance. 
As an example, the Hateful Memes Challenge and Multimedia Automatic Misogyny Identification (MAMI) benchmarks expose the inadequacy of unimodal systems, since signals stemming from text and image can be complementary and often individually ambiguous \cite{hateful-memes, fersini-etal-2022-semeval}. For this reason, the adoption of inherently multimodal approaches has been favored in related literature, with approaches ranging from feature-level text–image fusion to end-to-end vision-language models (VLMs) \cite{hakimov-etal-2022-tib, arango-etal-2022-hateu}.

The EXIST task series reflects the evolution of multimodal and perspectivist approaches to sexism identification, progressively moving from text-based sexism identification to disagreement-aware formulations that incorporate images, text, video and fine-grained metadata \cite{RodrguezSanchez2021OverviewOE, exist23, overview-exist23,exist24,exist25, exist26}. The 2026 edition, which constitutes the focus of the current paper, is centered around meme images and TikTok videos, accompanied by physiological and neurophysiological signals such as heart rate, eye tracking, and EEG.

In our approach, we contribute to the valuable field of multimodal sexism identification by proposing an efficient, multi-stage architecture grounded in late feature-level fusion. Rather than relying on a monolithic end-to-end vision-language transformer, our system leverages specialized modality encoders—including CLIP for visual semantics, multilingual transformers for OCR text, and acoustic descriptors for videos—alongside demographic and physiological metadata. A central component of our pipeline is the use of LLM-derived semantic indicators, which complement low-level visual and textual representations with higher-level cues aligned with the task taxonomy. %By prompting an LLM with combined visual-textual descriptions, we extract explicit, taxonomy-aligned indicators (such as stereotyping, objectification, and misogyny) that successfully capture nuanced, context-dependent sexism. 
By prompting an LLM with combined visual-textual descriptions, we extract explicit, taxonomy-aligned indicators, such as stereotyping, objectification, and misogyny, designed to capture higher-level semantic cues that are difficult to encode through low-level visual or lexical representations alone.
Finally, we implement a robust dimensionality reduction strategy coupled with a hierarchical XGBoost regression framework. %This ensures that our system effectively mitigates cross-modal noise—which proved crucial for the video modality—while strictly adhering to the soft-label, disagreement-aware constraints of the EXIST 2026 evaluation protocol. 
This allows the system to mitigate the impact of cross-modal noise while remaining compatible with the soft-label, disagreement-aware evaluation protocol of EXIST 2026.

\section{Related work}
\subsection{Sexism, Misogyny, and Hate Speech Detection}
Abusive language detection has extensively studied sexism, misogyny, and hate speech, drawing on psychological research. Prior work has shown that sexism is not limited to overt hostility, but is also  deployed in subtle and socially normalized forms including benevolent sexism, gender stereotyping, and objectification \cite{glick1996ambivalent, fredrickson1997objectification}. This indirect nature of gender-related discriminatory practices is reflected in their dissemination on social media, calling for computational approaches that do not primarily rely on the detection of explicit slurs or aggressive language, but may be concealed using humor, culturally situated references or implicit assumptions. At the same time, online misogyny has been recognized as a distinct form of gendered abuse, with digital platforms amplifying the visibility, reach, and persistence of hostile or discriminatory content targeting women \cite{misogyny, ijerph20105770}.

Several NLP benchmarks have addressed the aforementioned challenges over the years. The Automatic Misogyny Identification shared task \cite{Fersini2018OverviewOT} frames misogyny detection as a multilingual classification problem, harnessing tweets in English and Italian. In a similar sense, HatEval at SemEval-2019 \cite{basile-etal-2019-semeval} considers hate speech against women and immigrants harnessing English and Spanish Twitter data within a multilingual classification framework that also provides finer-grained signals such as aggressiveness and target type. As a parallel endeavor, the EXIST task series has progressively expanded to include source-intention detection, sexism categorization, multilingual settings, and disagreement-aware annotations \cite{RodrguezSanchez2021OverviewOE, overview-exist23, exist23}. With the advent of potent multimodal models, the incorporation of additional modalities becomes a timely concern, with later EXIST editions extending the task from tweets to memes and subsequently to TikTok videos \cite{exist24, exist25}. The trajectory continues this year with the integration of physiological and neurophysiological signals from annotators to provide a previously unexplored dimension connected to hateful and discriminatory online expressions \cite{exist26, exist26-extended}.

\subsection{Multimodal Sexism}
The growing prevalence of visually-rich social media discourse has motivated the attention to multimodal abusive-content detection. The interaction between visual, audio and linguistic elements may conceal the presence of explicit unimodal hateful signals, raising the need for jointly processing modalities. This need is further reinforced when complementary signals are not explicit or literal, but involve pragmatic context, irony, and implicit social stereotypes. To this end, several multimodal benchmarks have been introduced. The Hateful Memes Challenge was designed to require joint reasoning over image and text, including the use of benign confounders that make unimodal shortcuts unreliable \cite{hateful-memes}. MultiOFF similarly introduced a multimodal dataset for offensive meme detection, highlighting the need to combine textual and visual information in the analysis of online offensive content \cite{suryawanshi-etal-2020-multimodal}. Memotion Analysis further broadened the study of Internet memes by addressing sentiment, emotion, humor, sarcasm, offensiveness, and motivational content in a visuo-linguistic setting \cite{sharma-etal-2020-semeval}. Beyond classification, MOMENTA proposed a multimodal framework for detecting harmful memes and their targets, emphasizing both local and global perspectives in meme understanding \cite{pramanick-etal-2021-momenta-multimodal}. A more direct misogyny-related benchmark is Multimedia Automatic Misogyny Identification (MAMI) task at SemEval-2022 \cite{fersini-etal-2022-semeval}, which focused on detecting misogynous memes and identifying fine-grained misogyny categories such as shaming, stereotype, objectification, and violence. Evidence from participants showcased that multimodal fusion can improve the detection and characterization of misogynistic content, while also revealing the limitations of current systems when confronted with implicit meaning, visual stereotypes, and complex image–text interactions \cite{arango-etal-2022-hateu, hakimov-etal-2022-tib}.

Recent work has extended multimodal sexism detection beyond static images toward video-based and relational settings. Sexism identification on TikTok, integrating text, audio, and video features showcases that multimodal analysis can be particularly useful for understanding the intention behind sexist content \cite{tiktok}. MuSeD further contributes to this direction by providing a multimodal Spanish dataset for sexism detection in social-media videos \cite{grazia2025mused}. At the meme level, MemeWeaver proposes inter-meme graph reasoning for sexism and misogyny detection, moving beyond isolated instance-level classification by modeling relations among memes and their shared semantic patterns \cite{italiani-etal-2026-memeweaver}. 

\subsection{Learning with Disagreement}
Perceiving sexism and discrimination is often subjective and varies according to human perspectives and background. Disagreement offers a valuable signal within this socially sensitive and context-dependent setup, setting the cornerstone for multi-label handling instead of typical pipelines that resort to single-label majority-voting schemas. Learning with disagreements has therefore emerged as a learning paradigm that preserves subjectivity and varying human perspectives \cite{disagreement}. Systematic disagreement is valuable towards improving uncertainty estimation by modeling annotator-specific judgments \cite{davani-etal-2022-dealing}, especially when dealing with tasks that involve socially grounded, or value-laden judgments \cite{perspectivist}. This direction is formalized by evaluating models under soft-label and perspectivist paradigms, reflecting the distribution of human judgments in place of less informative, aggregated scores \cite{leonardelli-etal-2025-lewidi}. 

Disagreement-aware signals have been incorporated in recent EXIST series, expressing the inherent subjectivity of sexism among humans of different background, sensitivity to implicit stereotypes, interpretation of humor, or understanding of the communicative context \cite{exist23, exist24, exist25, exist26, exist26-extended}. In multimodal settings, disagreement may be further amplified because annotators can attend to different aspects of the input, such as the textual caption, the image, the speaker’s tone, or the broader cultural reference. For EXIST 2026, the inclusion of physiological and neurophysiological signals adds an additional human-centered layer to this problem, creating an opportunity to study not only what labels annotators assign, but also how users respond to sexist content at the perceptual and affective levels.

\section{Task Description}
EXIST 2026 focuses on the automatic identification and characterization of sexist content in complex multimedia social-media formats, specifically memes and TikTok videos, extending previous editions with sensor-based information (heart-rate-related measurements, eye tracking  and EEG) collected from users exposed to potentially sexist content. The dataset builds on previous EXIST resources for memes and TikTok videos, while enriching them with information reflecting both explicit annotator judgments and implicit user responses to the content.

The major advancement of EXIST 2026, i.e. the incorporation of sensor-based data, was performed in several stages, commencing from fitting the annotators with  sensors for ECG, eye tracking, and EEG, allowing a two-minute stimulus-free period to set a baseline. Consequently, they filled demographic  questionnaires to collect metadata including age, gender, education level, country of residence, and occupation. Finally, the annotators reported their daily social media usage and the percentage of time they consume in popular social media platforms. In the next stage of the experiments, individuals are tasked to rate their agreement regarding items in a six-point Likert scale, evaluating four cognitive dimensions (open thought, closed thought, intuitive thought, and effortful thought). The current emotional state of each participant was self-assessed before each experimental session. Between consecutive item assessments, a 3-second pause was inserted, in order to eliminate influences from the previous assessment, while concentration and comprehension of individuals was controlled using specified brief questions after each stimulus.

\subsection{Subtasks}
The shared task is organized around three main objectives: sexism identification, source-intention detection, and sexism categorization. Each objective is applied to two types of multimodal content: memes and TikTok videos; both of these modalities cover English and Spanish languages.

\textbf{Sexism identification} is a binary classification task in which systems must determine whether a given meme or video contains sexist expressions, serving as the entry point for identifying potentially harmful gender-related content and requires models to distinguish sexism from non-sexist material, including cases that may still be offensive, humorous, ironic, or controversial but not necessarily sexist.

\textbf{Source-intention detection} aims to characterize the intention behind the sexist message. This subtask is particularly challenging because sexist language or imagery may appear in different communicative contexts. For example, content may directly express sexist beliefs, report or criticize sexism, or use irony and quotation in ways that complicate the interpretation of the author’s stance. In this sense, source-intention detection goes beyond surface-level toxicity classification and requires models to infer the communicative function of the content.

\textbf{Sexism categorization} focuses on assigning fine-grained categories that describe the type of sexism expressed in the content, reflecting the fact that sexism is not a homogeneous phenomenon, but may involve ideological inequality, stereotyping and dominance, objectification, sexual violence, misogyny, or other forms of gender-based discrimination, depending on the official label taxonomy used in the task. Since a single instance may express multiple facets of sexism, this objective is typically more complex than binary identification and benefits from models capable of capturing subtle contextual and multimodal cues.

In the meme setting, systems receive image–text content, where sexist meaning may arise from the interaction between the visual scene and the overlaid or associated text. In the video setting, the input consists of TikTok videos and associated textual information, requiring models to account for dynamic visual content, speech or captions, and possibly audio-visual cues. In both settings, participants may also exploit the provided sensor-based signals, when available, as additional information about how users perceived or reacted to the content.

\subsection{Dataset}
\begin{figure}[h!]
    \centering
    \includegraphics[width=0.6\linewidth]{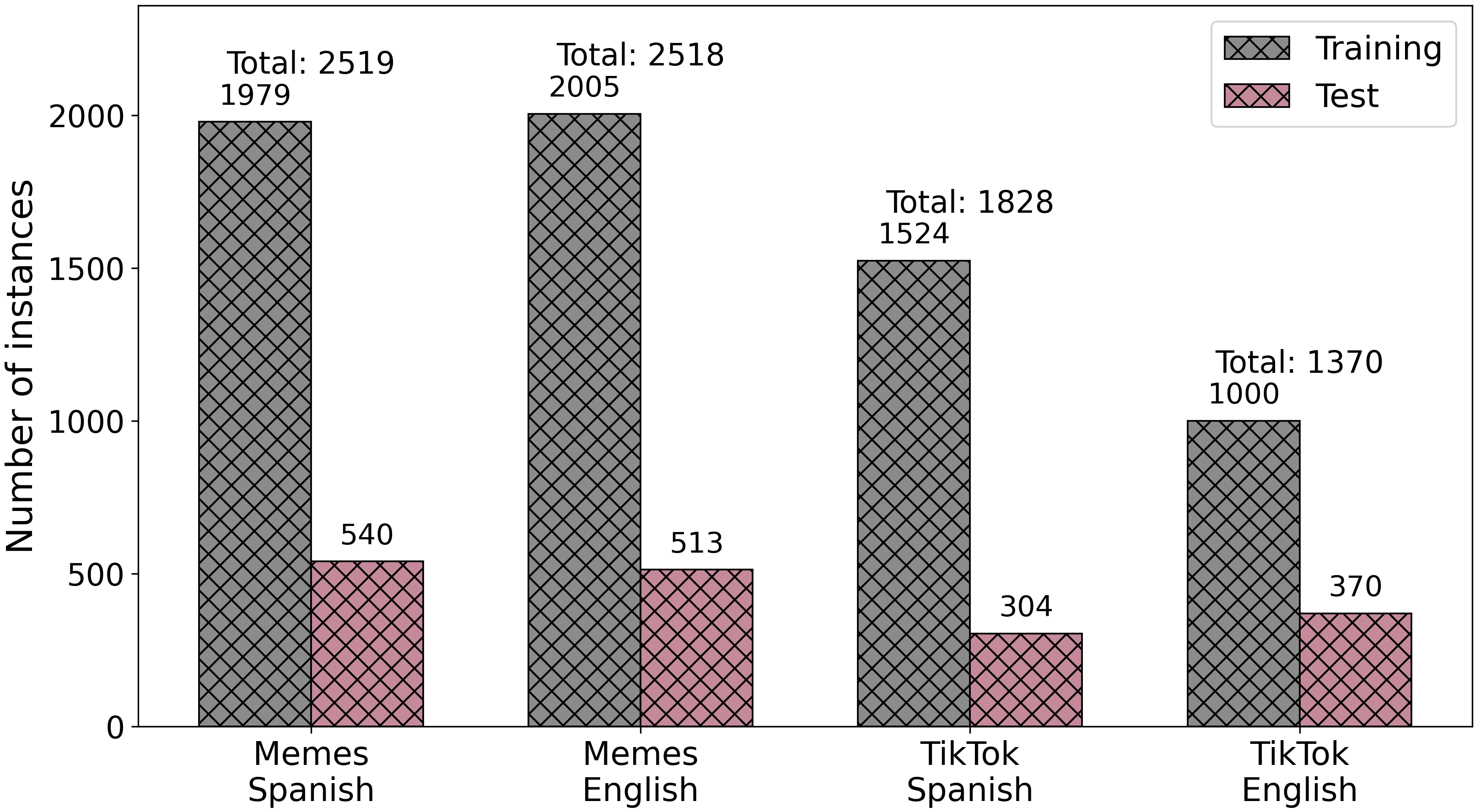}
    \caption{Official data statistics of EXIST 2026 \cite{exist26}.}
    \label{fig:stats}
\end{figure}

The current edition contains 8,235 multimedia instances in English and Spanish, with 5,037 of them corresponding to memes (3,984 training and 1,053 test memes), and  3,198 instances referring to TikTok videos (2,524 training and 674 test videos). Overall dataset statistics are presented in Figure \ref{fig:stats}.

To better understand the underlying data dynamics, it is crucial to examine the label distribution across the core subtasks. For the binary sexism identification (Tasks 2.1 and 3.1), the hard-label training ground truth provides explicit annotations for 3,370 meme instances (exhibiting a distribution of 2,003 \textbf{Yes} and 1,367 \textbf{No}) and 2,508 video instances. For the finer-grained tasks, the annotated subsets comprise 3,149 memes and 2,466 videos for source intention (Tasks 2.2 and 3.2), alongside 3,357 memes and 2,279 videos for multi-label categorization (Tasks 2.3 and 3.3). Regarding the multi-label categorization tasks, the dataset presents a natural long-tail distribution; broad socio-linguistic categories such as "Stereotyping" and "Ideological Inequality" appear much more frequently than extreme cases like "Sexual Violence". %This class imbalance inherently increases the difficulty of the prediction tasks, justifying our robust feature engineering and selection strategies.
This class imbalance increases the difficulty of the prediction tasks and motivates the use of soft-label modeling, feature selection, and category-aware analysis.

In alignment with the official evaluation protocol, our methodology explicitly accounts for the quality and consensus of the dataset's annotations. Specifically, instances marked as "UNKNOWN" in the ground truth due to a lack of annotator consensus (such as the unassigned training instances in Tasks 2.1 and 3.1) do not constitute a target prediction class. Therefore, they were systematically excluded from our training mechanisms to prevent the introduction of unresolvable noise into the gradient boosting classifiers.

\begin{figure*}[htbp]
    \centering
    \includegraphics[width=0.32\textwidth]{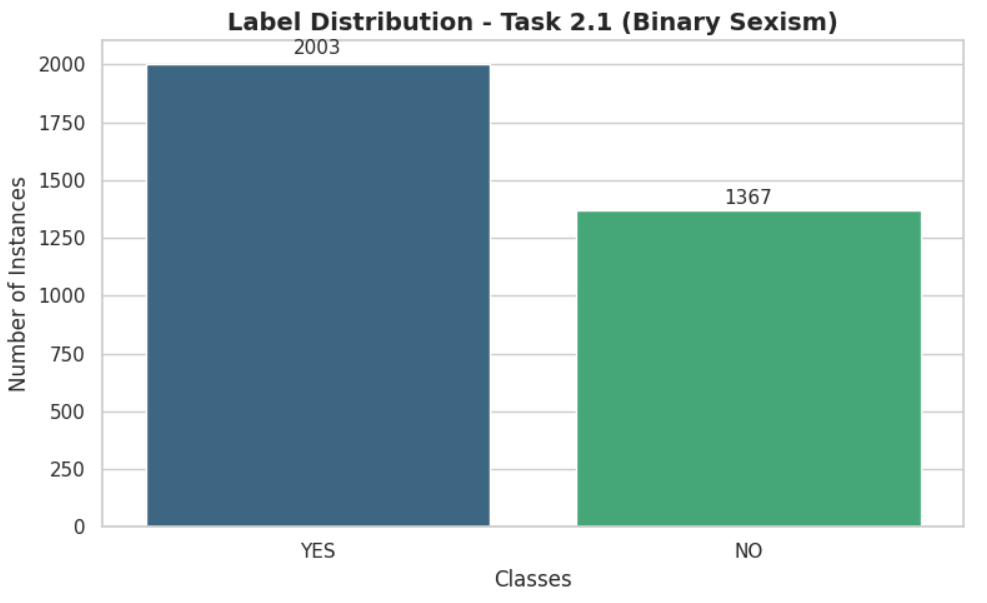} \hfill
    \includegraphics[width=0.32\textwidth]{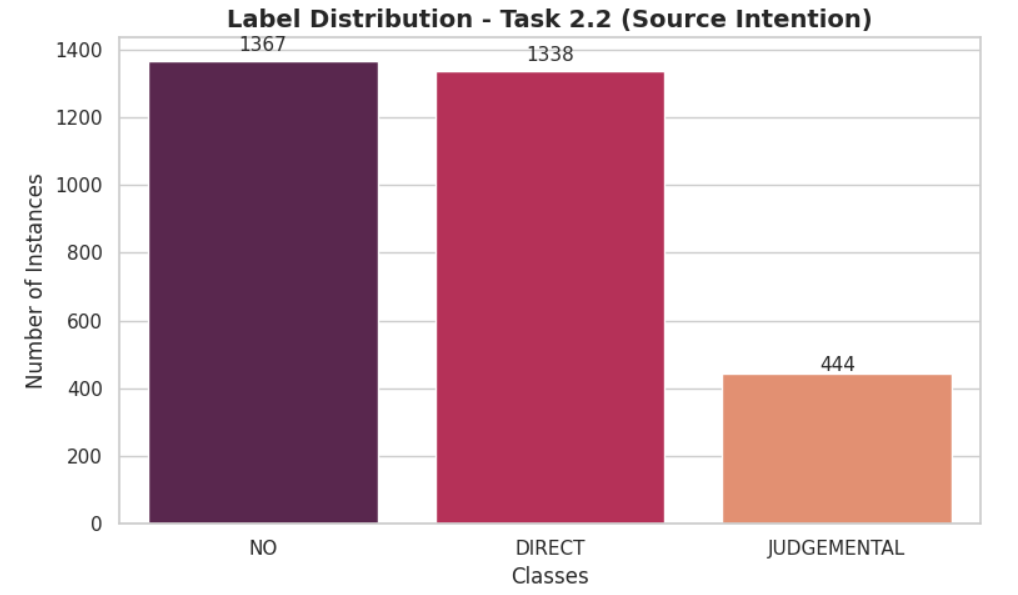} \hfill
    \includegraphics[width=0.32\textwidth]{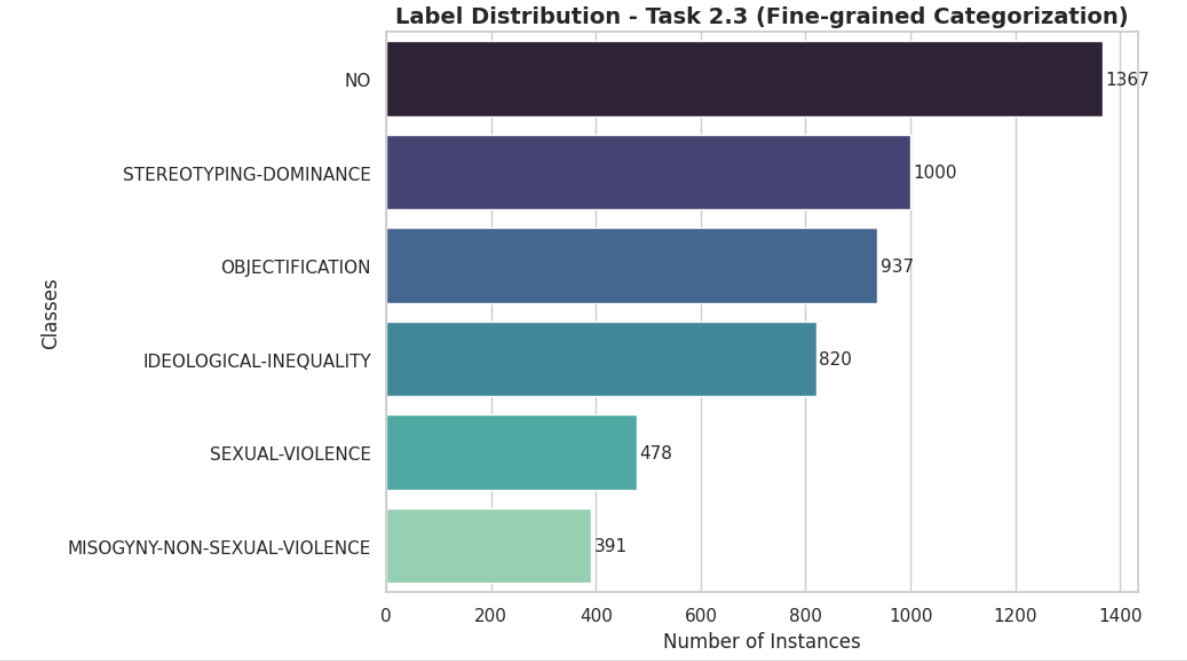}
    \caption{Label distribution across the Meme modality for the binary identification (Task 2.1), source intention (Task 2.2), and multi-label categorization (Task 2.3) subtasks in the training set.}
    \label{fig:meme_distributions}
\end{figure*}

\begin{figure*}[htbp]
    \centering
    \includegraphics[width=0.32\textwidth]{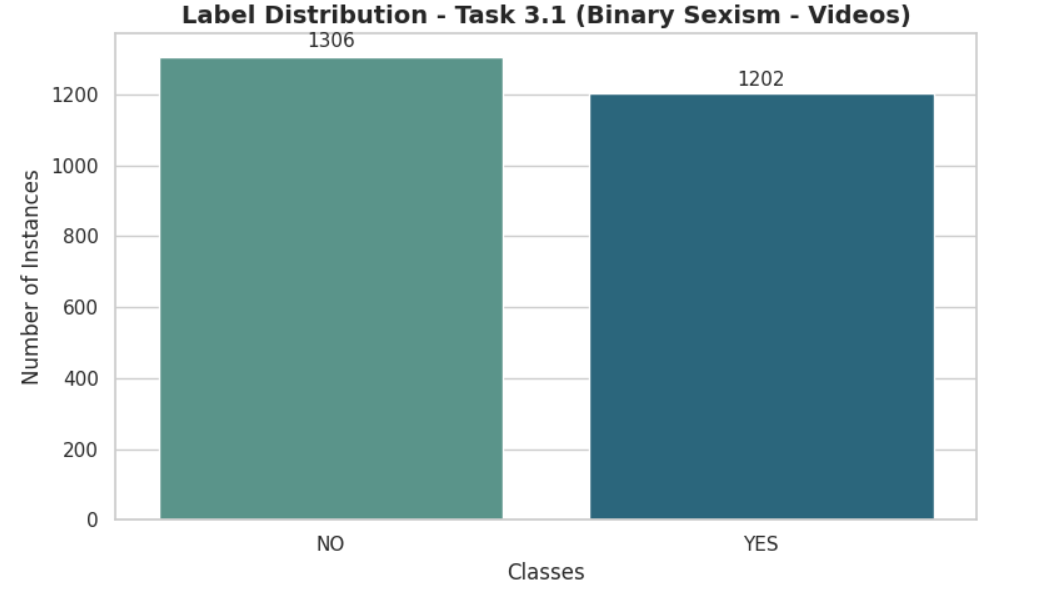} \hfill
    \includegraphics[width=0.32\textwidth]{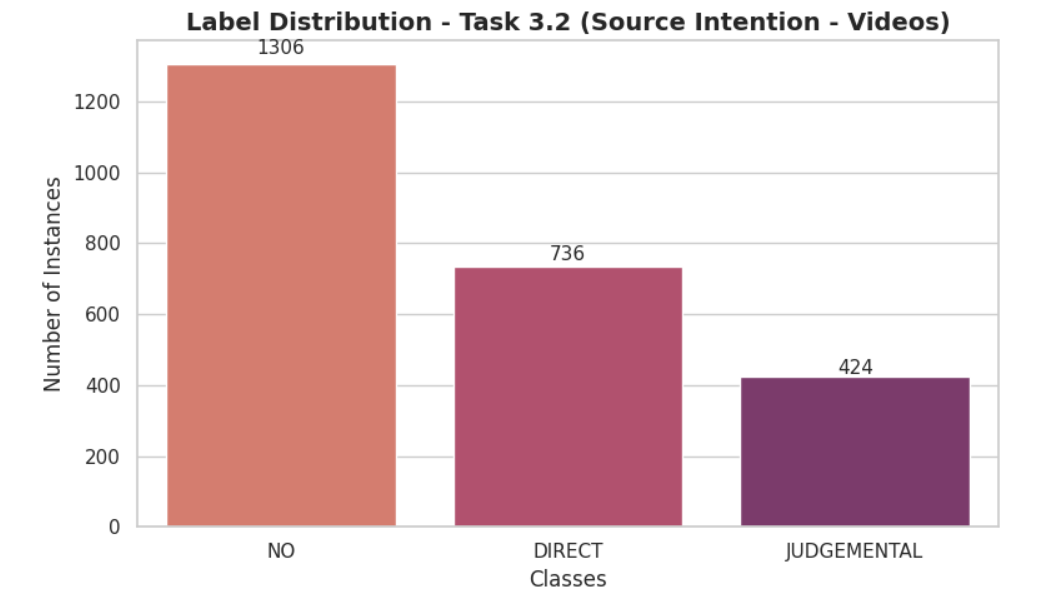} \hfill
    \includegraphics[width=0.32\textwidth]{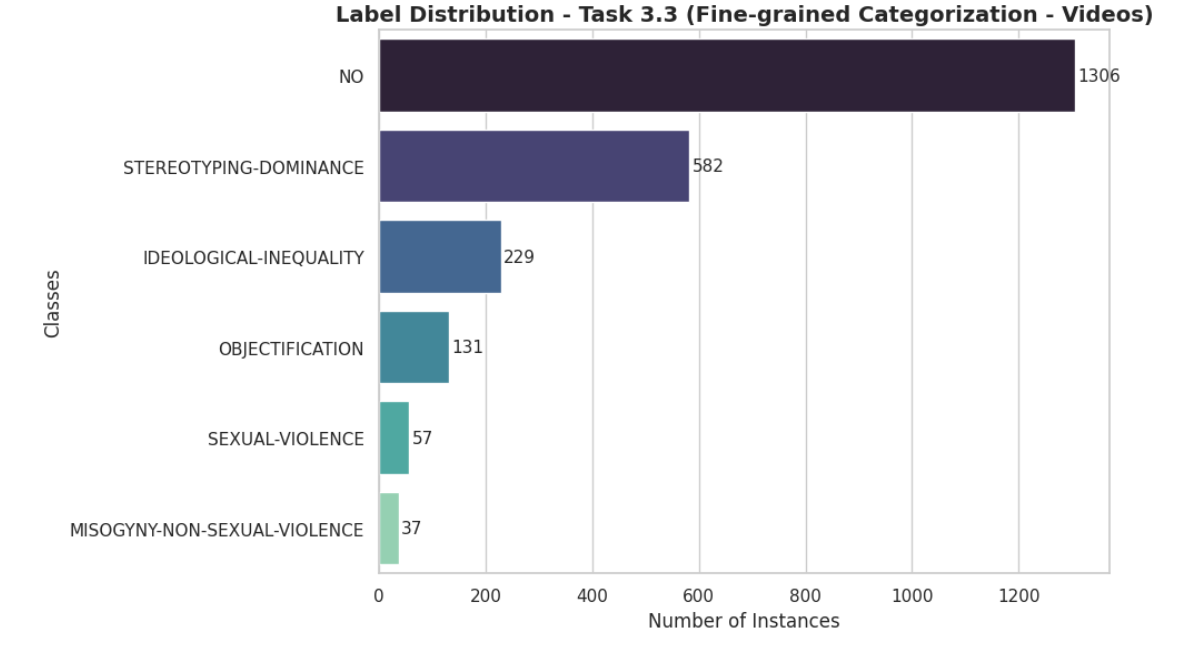}
    \caption{Label distribution across the Video modality for the binary identification (Task 3.1), source intention (Task 3.2), and multi-label categorization (Task 3.3) subtasks in the training set.}
    \label{fig:video_distributions}
\end{figure*}
\subsection{Evaluation}
EXIST 2026 follows the Learning with Disagreement evaluation protocol, leveraging two evaluation regimes: soft-soft and hard-hard. %Soft-soft regards probability distributions over valid labels instead of outputting a single  label. 
In the soft-soft setting, systems output probability distributions over valid labels rather than a single discrete label.
This setup reflects disagreement and uncertainty, and systems are evaluated via the \textit{ICM-Soft} metric \cite{amigo-delgado-2022-evaluating}, and secondarily using Cross-entropy. \textit{ICM} is an information-theoretic similarity metric for hierarchical multi-label classification, mathematically expressed as:
\begin{equation}
\mathrm{ICM}(s(d), g(d)) =
2\,IC(s(d)) + 2\,IC(g(d)) - 3\,IC(s(d) \cup g(d))
\end{equation}
where $s(d)$ refers to the set of labels predicted by the system, $g(d)$ is the gold annotation for an instance $d$ and $IC(\cdot)$ denotes information content. More generally, \textit{ICM} can be expressed as:
\begin{equation}
\mathrm{ICM}(A,B) =
\alpha_1 IC(A) + \alpha_2 IC(B) - \beta IC(A \cup B),
\end{equation}
where $A$ and $B$ are label sets and $\alpha_1$, $\alpha_2$, and $\beta$ are weighting parameters. The EXIST evaluation framework uses the instantiation $\alpha_1 = 2$, $\alpha_2 = 2$, and $\beta = 3$.

In the soft-soft setting, \textit{ICM-Soft} extends the same principle to soft label assignments, comparing the probability distribution predicted by the system with the empirical distribution of annotator judgments. For a category $c$ and agreement value $v$, the information content of a soft assignment is estimated as:
\begin{equation}
IC(\{\langle c, v \rangle\}) =
-\log_2 P\left(\{d \in D : g_c(d) \geq v\}\right),
\end{equation}
where $g_c(d)$ denotes the annotator-derived agreement level for category $c$ in instance $d$. This allows the evaluation to reward systems not only for predicting the majority label, but also for approximating the degree of annotator disagreement.

%On the other hand, systems that are evaluated on the hard-hard setting, outputting a single final label instead of label distributions utilize the \textit{ICM-Hard} metric. 
In the hard-hard setting, systems output a single discrete label and are evaluated using \textit{ICM-Hard}.
In this case, the unique gold label is extracted from enforcing a probabilistic threshold over annotators' labels. Complementarily, F1 score is  computed.
Both evaluation regimes are applicable on all subtasks for either memes or video data, following the valid labels per subtask. 

%\section{Dataset}
%\subsection{Overview}
%\subsection{Text Extraction and Normalization}
%\subsection{Visual and Video Processing}
%\subsection{Features}

\section{System Description}
\subsection{Architecture}
\label{sec:architecture}
\begin{figure}[t!]
    \centering
    \includegraphics[width=\linewidth, height=5.3cm]{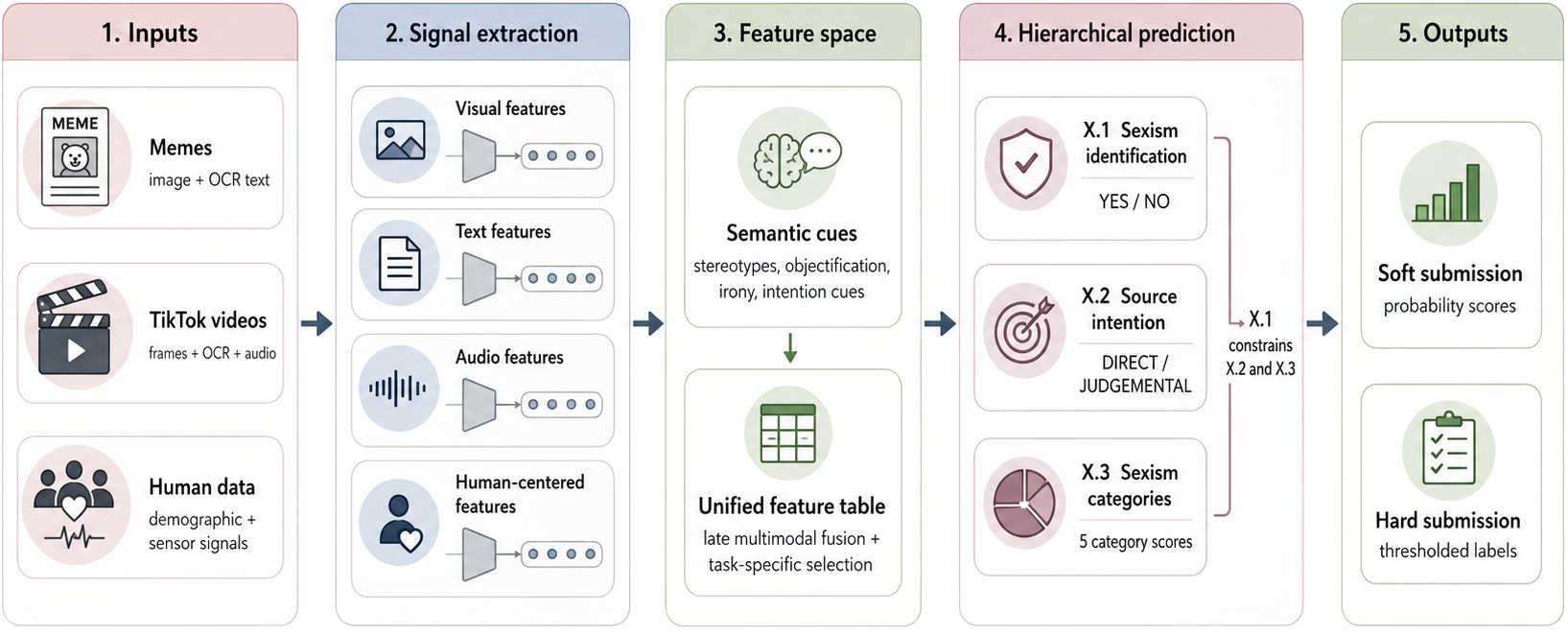}
    \caption{Hierarchical multimodal pipeline for EXIST 2026. Meme, video, and human-centered signals are converted into features, fused into a unified representation, and mapped to soft and hard predictions for sexism identification, source intention, and sexism categorization, covering 6 subtasks in total.}
    \label{fig:outline}
\end{figure}

Our system follows a modular, hierarchical feature-based multimodal architecture designed to transform heterogeneous information sources into task-specific soft predictions. Rather than relying on a single end-to-end vision-language transformer, our architecture decomposes the problem into consecutive stages: input processing, modality-specific representation extraction, semantic feature enrichment, feature-level fusion, task-specific prediction, and finally, output post-processing. This design was adopted to support both meme and video inputs, while allowing different types of evidence, including textual, visual, acoustic, demographic, biometric, and sensor-derived modalities to contribute to the final prediction. The system outline is illustrated in Figure \ref{fig:outline}.

\paragraph{Stage 1: Raw input processing.}The first stage receives the raw multimodal instance. For memes, the input consists of an image and its associated textual content, including text appearing on the image. For videos, the input consists of dynamic visual content, associated textual information, and audio-related signals. In both settings, the system may also exploit metadata and sensor-based information associated with the annotation process. These different sources are processed separately before being combined, so that each modality can contribute through a representation appropriate to its structure.

\paragraph{Stage 2: Signal extraction.}
In this stage, textual information is represented through multilingual text embeddings, while visual information is represented either through image/frame embeddings or through automatically generated image descriptions. For videos, additional acoustic representations are extracted from the audio stream in order to capture paralinguistic cues such as tone-related patterns. In this sense, the term “multimodal architecture” in our system does not refer to a single pretrained vision-language transformer used end-to-end. Instead, it refers to a pipeline that integrates representations derived from multiple modalities into a unified feature space.

\paragraph{Stage 3: Feature enrichment.}
This stage enriches the feature space with higher-level semantic indicators. As sexism in memes and short videos may depend on implicit stereotypes, irony, objectification, or authorial stance, low-level embeddings alone may not capture all relevant signals. We therefore add compact semantic features derived from language-model prompting over combined visual-textual descriptions. These features are designed to make explicit certain interpretable properties of the content, e.g. whether it contains stereotypes, objectification, irony, or category-specific sexism cues.

\paragraph{Stage 4: Feature-level fusion.}
In this stage, all available representations—textual, visual, acoustic, demographic, biometric, sensor-derived, and LLM-derived semantic indicators—are concatenated into a structured feature table. This produces a unified representation for each instance, while preserving the possibility of selecting different feature subsets for different subtasks or modalities. This late-fusion design allows the system to remain flexible: components can be added or removed without retraining a large end-to-end multimodal model.

\paragraph{Stage 5: Output.}
The final prediction stage maps the fused feature representation to soft scores for each  subtask. For binary sexism identification, the system estimates the probability that the instance is sexist (\textbf{Yes} or \textbf{No}). For source-intention detection, it estimates scores over intention labels (\textbf{Direct} or \textbf{Judgemental}). For sexism categorization, it estimates scores for the fine-grained sexism categories  (\textbf{Ideological  and Inequality}, \textbf{Stereotyping and Dominance}, \textbf{Objectification}, \textbf{Sexual Violence}, and \textbf{Misogyny and Non-Sexual Violence}). These soft outputs are submitted directly in the soft evaluation setting. For the hard evaluation setting, they are converted into discrete labels using  thresholding rules: for sexism identification, instances were labeled as \textbf{Yes} when the predicted sexism probability exceeded 0.5, and as \textbf{No} otherwise. For source-intention detection, the final label was selected using the maximum predicted score between \textbf{Direct} and \textbf{Judgemental}. For sexism categorization, a multi-label thresholding strategy was applied where all categories with a predicted probability of at least 0.33 were selected, reflecting the intuition that a category should be retained when supported by more than one annotator. 
%To ensure robust predictions in edge cases where no specific sexism category reached this threshold, a fallback mechanism was implemented: if the probability of the "NO" (non-sexist) class was 0.5 or higher, the instance was assigned the "NO" label; otherwise, the instance was assigned the single sexism category that yielded the maximum predicted probability (argmax) 

\begin{figure}[h!]
    \centering
    \includegraphics[width=0.75\linewidth]{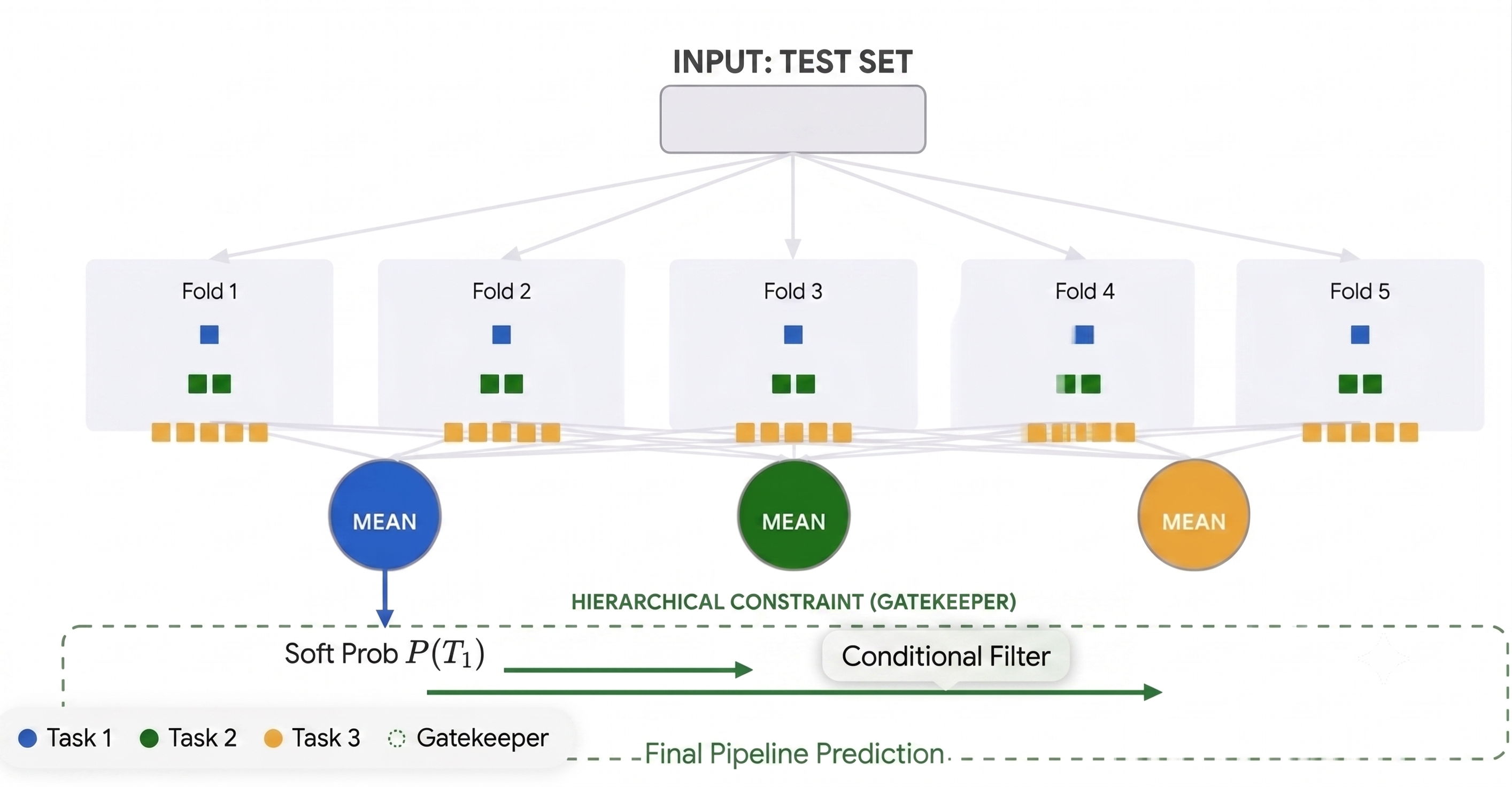}
    \caption{Overview of the hierarchical multimodal ensemble architecture. The system employs a 5-fold cross-validation strategy, training independent XGBoost regressors for each target label across subtasks (40 models in total). During inference, fold-specific predictions are averaged to produce robust soft labels, which are subsequently bounded by the hierarchical normalization layer with Task X.1 acting as a probabilistic gatekeeper. }
    \label{fig:architecture}
\end{figure}
\subsection{Demographic and Biometric Feature Engineering}
To effectively incorporate the subjective perspectives of the annotators into our prediction pipeline, we engineered a robust set of continuous metadata features. Rather than treating annotator demographics—such as gender, age group, ethnicity, educational level, and country of origin—as discrete categorical variables, we transformed them into instance-level frequency distributions. Specifically, for each meme or video, we calculated the ratio of each demographic category present among its annotators. This yielded normalized continuous features representing the demographic composition of the annotator pool per instance. Missing demographic values were imputed with zeros.

Furthermore, physiological and sensor data (Eye-Tracking, Heart Rate, and EEG) were aggregated to form a unified instance-level representation. For items with available sensor readings, we computed the arithmetic mean of the signals across all participating users. Recognizing that sensor data is frequently sparse, we imputed missing physiological values with zeros while simultaneously introducing explicit binary missing-indicator flags (\texttt{is\_missing}) for every sensor feature. This structural design allows the gradient boosting models to mathematically distinguish between a true zero-magnitude physiological response and a complete absence of sensor data.
\subsection{Multimodal Fusion and Hierarchical Normalization}
\label{sec:fusion}
Our system relies primarily on late feature-level fusion. Instead of training a single end-to-end multimodal transformer, we extract representations from each available information source and concatenate them into a unified feature vector. This design allows the available heterogeneous signals to be jointly exploited by the downstream predictors. For memes, the fused representation combines textual and visual information in two complementary ways. First, the meme image is represented through visual embeddings, while the text written on the image is represented through multilingual textual embeddings. Second, the image is converted into a natural-language description and concatenated with the OCR text, producing a multimodal textual representation that captures both visual and linguistic content. This representation is further used to derive LLM-based semantic indicators, such as stereotype, objectification, irony, or misogyny-related cues. For videos, the fused representation includes frame-level visual features, OCR-derived textual embeddings, acoustic features, demographic metadata, and biometric or sensor-derived information. The video pipeline therefore integrates static visual cues, textual cues, and audio-related descriptors, while keeping all modalities in a common tabular representation suitable for downstream prediction.

Beyond feature-level fusion, we also apply a hierarchical normalization strategy that reflects the dependency among subtasks. Sexism identification acts as a \textit{gatekeeper}: the predicted probability of the \textbf{Yes} label in the binary subtasks constrains the downstream predictions for source intention and sexism categorization. For source-intention detection, the predicted scores for \textbf{Direct} and \textbf{Judgemental} are normalized so that their sum matches the predicted sexism probability. For sexism categorization, the scores of the five fine-grained categories are clipped so that no category probability exceeds the predicted probability of sexism. This prevents inconsistent outputs, such as assigning high confidence to a sexism category while predicting that the instance is non-sexist.

\subsection{Feature Selection}
Feature selection is a task-specific component of our pipeline, implemented to control the dimensionality and noise introduced by the heterogeneous multimodal representation. Since the fused feature space combines signals from different sources, not all dimensions are expected to be equally informative for every subtask. Applying feature selection allows the system to retain the most informative signals for each prediction problem while reducing the influence of weak, redundant, or noisy features.

%To operationalize this, we employ a tree-based feature selection methodology driven by the intrinsic evaluation metrics of gradient boosting. 
To operationalize feature selection, we use a tree-based ranking procedure. An initial XGBoost regressor, configured with 200 estimators and histogram-based tree growth, is trained on the unreduced feature space. We then extract the feature importances from the fitted model, rank all dimensions in descending order, and retain the top-ranked features for downstream training. In development experiments, we compare different cardinality thresholds and find that retaining 300 features provides the best trade-off between information preservation and noise reduction for meme Task 2.1.

In addition to improving computational efficiency and robustness against cross-modal noise, this selection provides a vital interpretability mechanism. By inspecting the top-ranked dimensions, we can better understand which types of evidence the system relies on, such as textual cues, visual representations, sensor-derived information, or higher-level semantic indicators. This is particularly important in sexism detection, where model decisions may depend on subtle cues related to stereotypes, objectification, irony, or authorial stance.

\section{Experiments}
% Mention about experimental setup and environment
\subsection{Models}
For meme inputs, we extract three main types of representations. First, visual content is represented directly through CLIP \cite{clip} image embeddings. Second, each image is converted into a natural-language description using BLIP \cite{blip} as the captioner; this description is concatenated with the OCR text and encoded with XLM-RoBERTa \cite{xlmr}, producing a multimodal textual representation. Third, the same BLIP–OCR textual input is passed to Qwen2.5-1.5B-Instruct \cite{qwen2_5} using constrained prompts that returned binary semantic indicators. These indicators capture higher-level cues such as irony, stereotyping, objectification, source intention, and category-specific sexism signals.

For video inputs, we leverage a parallel feature-based representation. Visual information is extracted from video frames using CLIP, textual information is encoded from OCR text using paraphrase-multilingual-MiniLM-L12-v2 from SBERT \cite{reimers-2019-sentence-bert}, and acoustic information is represented through MFCC features extracted with librosa \cite{librosa}. Demographic, biometric, and sensor-derived features are included as additional tabular signals. We also tested variants with Whisper-based \cite{radford2022whisper} audio transcription and multiple CLIP frames, but these variants did not improve the development performance and were not selected as the primary submitted configuration.

%The final prediction layer was implemented using Extreme Gradient Boosting (XGBoost) regressors. Specifically, rather than standard squared-error regression, the models were configured with the \verb|reg:logistic| objective function. This ensures that the trees are trained to directly output bounded, continuous probability estimates in the \$[0, 1]\$ range, making them highly suitable for our soft-label targets. Furthermore, the training process utilized histogram-based tree growth (\verb|tree_method='hist'|) coupled with L1 and L2 regularization to mitigate overfitting across the heavily dimensional feature space. The raw probability outputs produced by these models were subsequently normalized or clipped according to the hierarchical strategy described in Section  \ref{sec:fusion}.
The final prediction layer was implemented using XGBoost regressors, which output continuous soft scores for the valid labels of each subtask. We used the \verb|reg:logistic| objective so that predictions are bounded in the $[0,1]$ range, making them suitable for the soft-label setting of EXIST. The resulting outputs were subsequently normalized or clipped according to the hierarchical strategy described in Section~\ref{sec:fusion}.

\subsection{Training, Hyperparameter Selection, and Ensembling}

All main predictors are trained as soft-label regression models, with model selection driven by development-time ICM-Soft. Since official test labels are unavailable during system development, we use 5-fold stratified cross-validation: each model is trained on four folds and validated on the remaining fold. Hyperparameters are optimized with Optuna \cite{akiba2019optuna} using 30 trials per configuration, with ICM-Soft as the validation objective.

For efficient training over high-dimensional feature spaces, we use histogram-based tree growth (\verb|tree_method='hist'|). L1 and L2 regularization terms are included in the hyperparameter search to reduce overfitting. Feature selection is performed before training the final task-specific predictors, allowing each subtask to rely on a compact subset of the fused representation.

The multi-task setting is implemented through label-wise decomposition. For Task X.1, one regressor is trained per fold to predict the probability of the \textbf{Yes} label. For Task X.2, two independent regressors are trained per fold, one for each source-intention label. For Task X.3, five independent regressors are trained per fold, one for each sexism category. Thus, each full configuration requires forty fold-specific regressors: five for Task X.1, ten for Task X.2, and twenty-five for Task X.3.

At test time, each official test instance is evaluated by the five fold-specific models for each target label, and the final soft score is computed as the arithmetic mean of the five predictions. The averaged soft predictions are then passed to the hierarchical normalization layer described in Section~\ref{sec:fusion} and depicted in Figure \ref{fig:architecture}. For hard-label submissions, soft outputs are converted using task-specific rules: thresholding at 0.5 for sexism identification, argmax for source intention, and thresholding at 0.33 for sexism categorization.

\subsection{Baselines and Ablations}

We organize the experimental variants as controlled modifications of the feature-based pipeline. Each experiment is defined by the operation it performs over the reference configuration: adding a new information source, removing a component, substituting a modeling strategy, or selecting a compact subset of features. To avoid ambiguity, we refer to the meme experiments as E1--E8 (Table \ref{tab:meme_experiments}) and the video experiments as E9--E14 (Table \ref{tab:video_experiments}).

\begin{table}[h!]
\centering \small
\caption{Meme experimental variants. Each experiment modifies the previous feature-based pipeline by \textbf{adding}, \textbf{removing}, \textbf{substituting}, or \textbf{selecting} components.}
\label{tab:meme_experiments}
\small
\begin{tabular}{p{0.02\linewidth} p{0.1\linewidth} p{0.40\linewidth} p{0.38\linewidth}}
\toprule
ID & Operation & Configuration & Purpose \\
\midrule
E1 & \opbase & Late-fusion meme baseline: demographic features, biometric/sensor features, and BLIP--OCR text embeddings. & Establish the initial feature-based multimodal baseline. \\

E2 & \opadd & E1 + CLIP image embeddings. & Test whether direct visual embeddings improve over the initial baseline. \\

E3 & \opsel & E2 with top-ranked feature subset. & Test whether feature selection reduces noise in the CLIP-enriched representation. \\

E4 & \oprem & E2 without BLIP descriptions; encode only OCR text while keeping the remaining feature structure.& Test whether generated image descriptions add useful context beyond meme text. \\

E5 & \opsub & Replace late feature-level fusion with ViLT-based early image--text fusion. & Test whether direct image--text interaction modeling improves over tabular late fusion. \\

E6 & \opadd & E2 + three compact semantic indicators: irony, male-directed insult, and female stereotype. & Test whether a small set of pragmatic/stereotype cues improves prediction. \\

E7 & \opadd & E2 + ten taxonomy-aligned semantic indicators covering source intention and sexism categories. & Test broad semantic enrichment aligned with the official label space. \\

E8 & \opadd & E2 + three focused semantic indicators: stereotyping/dominance, objectification, and misogyny/non-sexual violence. & Test whether compact sexism-specific cues outperform broader semantic prompting. \\
\bottomrule
\end{tabular}
\end{table}

\begin{table}[h!]
\centering \small
\caption{Video experimental variants. The video experiments test feature selection, additional speech/visual information, semantic indicators, and an alternative prediction model.}
\label{tab:video_experiments}
\small
\begin{tabular}{p{0.02\linewidth} p{0.1\linewidth} p{0.40\linewidth} p{0.38\linewidth}}
\toprule
ID & Operation & Configuration & Purpose \\
\midrule
E9 & \opbase & Full video feature baseline: demographic features, biometric/sensor features, CLIP frame embeddings, OCR text embeddings, and MFCC acoustic descriptors. & Establish the no-feature-selection video baseline. \\

E10 & \opsel & E9 with top-k selected features (k=300 and k=400). & Test whether a compact feature subset improves generalization. \\

%E11 & \opsel & E9 with top 400 selected features. & Test whether a broader selected subset improves over the top-300 setting. \\

E11 & \opadd & E10 + Whisper-based speech transcription and use multiple CLIP frames.& Test whether richer audio-text and visual coverage improves video prediction. \\

E12 & \oprem & E11 without Whisper transcription while retaining the visual/textual video feature structure.& Test whether automatic speech transcription introduces noise. \\

E13 & \opadd & E10 + LLM-derived semantic indicators analogous to those used for memes.& Test whether meme-style semantic cues transfer to video subtasks. \\

E14 & \opsub & Replace the main gradient-boosted predictor with CatBoost. & Test sensitivity to the prediction model family. \\
\bottomrule
\end{tabular}
\end{table}

The distinction between meme and video experiments reflects the different structure of the two modalities. For memes, the ablations primarily examine the contribution of visual embeddings, generated image descriptions, early versus late fusion, and LLM-derived semantic cues. For videos, the ablations focus on feature selection, speech transcription, multi-frame visual information, semantic indicators, and model-family substitution. The development results of these experiments are reported in Section~\ref{sec:dev_results}.

\section{Results}

\subsection{Development Results and Ablation Studies}
\label{sec:dev_results}
\paragraph{Memes.}
%Table \ref{tab:meme_results} summarizes the main development results for the meme subtasks in terms of the ICM-Soft metric. The initial late-fusion baseline (E1) obtained negative scores (-0.6577). Adding CLIP image embeddings (E2) substantially improved Task 2.1, reducing the error margin to -0.4919. An initial attempt to explicitly model irony via a cosine-based sarcasm score degraded the performance to -0.5145, indicating that low-level distance metrics were insufficient. 

%Regarding feature selection (E3), evaluating different subsets revealed that retaining the top 300 features yielded an optimal ICM-Soft of -0.4734, outperforming a stricter threshold of top 100 (-0.5117). Consequently, all subsequent meme experiments (E4 through E8) were trained exclusively on this top 300 feature space. Removing the BLIP generated descriptions completely (E4) slightly degraded the selected representation (-0.4760). Conversely, early fusion using ViLT (E5) severely failed, yielding -0.7838 on the full space and -0.7107 after selection, confirming the stability of our tabular late-fusion approach. The best meme sexism-identification result was obtained by the focused semantic feature set (E8), proving that compact, task-relevant LLM cues (stereotyping, objectification, and misogyny) successfully guide the gradient boosting predictors.

Table \ref{tab:meme_results} summarizes the main development results for the meme subtasks in terms of ICM-Soft. The initial late-fusion baseline (E1) obtains a score of -0.6577 on Task 2.1. Adding CLIP image embeddings (E2) substantially improves performance, reaching -0.4919. This suggests that the original BLIP--OCR textual representation does not fully capture visually encoded cues that are central in meme-based sexism, such as gendered visual stereotypes, objectifying imagery, or the visual setup required to interpret the overlaid text. The improvement obtained by CLIP therefore indicates that direct visual embeddings provide complementary information beyond generated image descriptions and OCR text.

An initial attempt to explicitly model irony through a cosine-based sarcasm score degrades ICM-Soft performance to -0.5145. This result suggests that irony in sexist memes is not adequately captured by a simple distance-based mismatch between image and text embeddings. In memes, ironic or sarcastic meaning often depends on pragmatic framing, cultural references, template conventions, and the intended stance of the author. A low-level similarity or dissimilarity score is therefore too coarse to distinguish between genuinely contradictory image--text pairs, humorous but non-sexist content, and sexist content expressed through irony.

Feature selection (E3) further improves the CLIP-enriched representation. Retaining the top 300 features yields the best ICM-Soft score of -0.4734, whereas a stricter top-100 subset degrades performance to -0.5117. This indicates that the fused representation contains a \textbf{mixture of informative, redundant, and noisy dimensions}. A moderate feature subset removes part of the noise introduced by heterogeneous sources, while still preserving enough multimodal information to capture diverse sexism cues. In contrast, the top-100 setting appears too restrictive, discarding useful complementary signals. Consequently, subsequent meme experiments (E4--E8) are trained on the top-300 feature space.

Removing BLIP-generated descriptions (E4) slightly deteriorates the selected representation, reaching -0.4760. The small magnitude of this drop suggests that BLIP descriptions provide useful but not decisive information. This is plausible because image captions can summarize salient visual content, but may omit socially relevant details such as gender roles, sexualization, facial expressions, or meme-specific contextual cues, especially since the underlying captioner is not explicitly trained on such context. Thus, BLIP contributes complementary visual semantics, but direct CLIP embeddings and OCR-based information remain the dominant signals in this configuration.

The early-fusion ViLT alternative (E5) performs substantially worse, with -0.7838 on the full feature space and -0.7107 after feature selection. Although early fusion is theoretically attractive for image--text reasoning, this result suggests that the available training setting is not sufficient for the model to learn robust cross-modal interactions. Meme-based sexism is highly contextual, noisy, and often implicit; under these conditions, a tabular late-fusion pipeline using pretrained representations appears more stable than an early-fusion vision-language model. This also supports the design choice of treating multimodal encoders as feature extractors rather than fine-tuning a single end-to-end multimodal architecture.

The strongest Task 2.1 result is obtained by the focused semantic configuration (E8). This finding highlights the value of compact, task-aligned LLM-derived indicators. Unlike broad semantic augmentation, the focused feature set targets core manifestations of sexism, namely stereotyping, objectification, and misogyny/non-sexual violence. These concepts are directly aligned with the phenomena that distinguish sexist from non-sexist memes, and they provide the gradient boosting models with high-level cues that are difficult to infer from embeddings alone. The result also suggests that LLM-derived features are \textbf{most effective when they are constrained}, interpretable, and closely tied to the target taxonomy, rather than when they attempt to encode many loosely related properties. The complete prompt templates, including the exact questions and formatting constraints provided to the LLMs for feature extraction, are detailed in Appendix \ref{appendix:prompts}.

Applying the hierarchical normalization strategy, we evaluate different combinations of gatekeeper models and downstream predictors for Tasks 2.2 and 2.3. For source-intention detection (Task 2.2), using the focused semantic configuration (E8) for both the binary gatekeeper and the downstream classifier produces the best development score of -1.2621. This suggests that source intention \textbf{benefits from a strong and semantically focused estimate} of whether the content is sexist in the first place. If the gatekeeper captures core sexist cues such as stereotyping and objectification, the downstream intention classifier receives a more reliable probability mass to distribute between \textbf{Direct} and \textbf{Judgemental}.

For sexism categorization (Task 2.3), the best score of -5.4329 is obtained by combining the comprehensive taxonomy-aligned predictors (E7) with the pragmatic gatekeeper model (E6). This difference is informative: binary sexism identification and fine-grained categorization do not require exactly the same information. The pragmatic gatekeeper appears to provide a \textbf{robust estimate of overall sexism presence}, while the broader E7 semantic features are better suited for \textbf{distinguishing among specific sexism categories}. In other words, compact semantic cues are most effective for deciding whether sexism is present, whereas taxonomy-aligned cues become more useful once the model must allocate probability mass across fine-grained categories.

\begin{table}[!t]
\centering
\caption{Development results for Meme Sexism Identification (Task 2.1). The table reports the ICM-Soft score for the optimal variant of each ablation step corresponding to the configurations defined in Table \ref{tab:meme_experiments}. Best values are denoted with \textbf{bold}.}
\label{tab:meme_results}
\begin{tabular}{llc}
\hline
\textbf{ID} & \textbf{Configuration} & \textbf{ICM-Soft $\uparrow$ (Task 2.1)} \\
\hline
\textbf{E1} & BASE (Demographics, Biometrics, BLIP-OCR) & -0.6577 \\
\textbf{E2} & ADD CLIP image embeddings & -0.4919 \\
\textbf{E3} & SELECT Top-$k$ features ($k=300$) & -0.4734 \\
\textbf{E4} & REMOVE BLIP descriptions & -0.4760 \\
\textbf{E5} & SUBSTITUTE ViLT early fusion & -0.7107 \\
\textbf{E6} & ADD 3 compact pragmatic cues & -0.4581 \\
\textbf{E7} & ADD 10 taxonomy-aligned cues & -0.5046 \\
\textbf{E8} & \textbf{ADD 3 focused sexism cues }& \textbf{-0.4482 (Best)} \\
\hline
\end{tabular}
\end{table}

%Applying our hierarchical normalization strategy, we evaluated different combinations of gatekeeper models and downstream predictors to formulate our final submissions. For source intention (Task 2.2), utilizing the focused semantic configuration (E8) for both the binary gatekeeper and the downstream classifier yielded the optimal development score of \textbf{-1.2621}. For sexism categorization (Task 2.3), the best development score of \textbf{-5.4329} was achieved by combining the comprehensive taxonomy-aligned predictions (E7) constrained by the pragmatic gatekeeper model (E6).

\paragraph{Videos.}
%Table \ref{tab:video_results} presents the corresponding development results for the video modality. Videos proved highly sensitive to cross-modal noise. The full unselected baseline (E9) yielded -0.0032. For the video modality, aggressive dimensionality reduction (E10) was critical: selecting the top 300 features achieved the best score (0.0378), whereas a broader subset of top 400 degraded performance (-0.0656). Therefore, the optimal top 300 subset served as the foundation for the subsequent experiments (E13, E14).

%Expanding the visual and audio context (E11) by including Whisper transcriptions and 3-frame CLIP embeddings introduced noise, dropping the score to -0.0200. Removing Whisper while keeping the 3 frames (E12) also failed to surpass the single-frame SOTA, even after feature selection (yielding 0.0159 at $k=400$). Interestingly, unlike memes, adding LLM-derived semantic indicators (E13) to the video pipeline yielded marginal improvements (0.0013) because these features were frequently filtered out by the XGBoost importance criteria. Finally, substituting the regressor with CatBoost (E14) proved highly unstable in this sparse space (-0.3588).
Table \ref{tab:video_results} presents the corresponding development results for the video modality. Compared with memes, videos prove more sensitive to cross-modal noise and feature dimensionality. The full unselected baseline (E9) obtains an ICM-Soft score of -0.0032 for Task 3.1, indicating that the initial multimodal representation already provides a relatively competitive starting point. However, the best result is achieved after aggressive feature selection: selecting the top 300 features (E10) improves the score to 0.0378, whereas a broader top-400 subset degrades performance to -0.0656. This suggests that the \textbf{video feature space contains many weak or noisy dimensions}, and that a compact subset is better suited for generalization. In contrast to memes, where several complementary semantic and visual signals improve performance, the video modality appears to benefit more from suppressing irrelevant features than from expanding the representation.

Expanding the visual and audio context (E11) by adding Whisper-based transcriptions and CLIP embeddings from three frames does not improve performance, yielding -0.0200. This result suggests that additional automatically extracted video information \textbf{may introduce noise} when it is not sufficiently aligned with the target labels. Whisper transcriptions \textbf{can be useful when speech carries the main discriminatory signal}, but they may also introduce transcription errors, irrelevant spoken content, or redundant information already partially captured by OCR. Similarly, using multiple frames increases visual coverage, but may dilute the signal if the selected frames contain background, transitions, or visually irrelevant moments. Since TikTok videos are temporally dynamic, a small number of sampled frames may not reliably capture the moment where sexist meaning is expressed.

%Removing Whisper while keeping the multi-frame setting (E12) yields a baseline score of -0.0319 using the full feature set, which significantly improves to 0.0159 after applying feature selection at k=400 , but still does not surpass the single-frame top-300 configuration. 
%This indicates that the speech-transcription component is not the only source of noise; the broader multi-frame representation itself may have introduced additional irrelevant visual variation. 
The E12 result further indicates that the speech-transcription component is not the only source of noise: even after removing Whisper, the multi-frame representation remains below the single-frame top-300 configuration. 
The stronger performance of the single-frame setup suggests that, in this pipeline, a compact representative visual snapshot combined with OCR, acoustic, demographic, and biometric features is more stable than a richer but noisier temporal representation.

Unlike in the meme setting, adding LLM-derived semantic indicators to the video pipeline (E13) produces only a marginal score of 0.0013. This difference is informative: the LLM features are generated using a strategy originally designed for memes, where a static image description and OCR text can be summarized into a compact visual-textual prompt. Videos, however, contain temporal, acoustic, and contextual information that is not fully captured by a static description or OCR text alone. As a result, meme-style semantic indicators such as irony, stereotypes, or objectification \textbf{may fail to reflect the relevant video-level meaning}. Their low feature importance further suggests that the downstream model did not find them reliable enough to retain as predictive signals.

Finally, substituting the predictor with CatBoost (E14) substantially degrades performance, reaching -0.3588. This suggests that, under the current feature representation and validation setup, the original XGBoost-based regression pipeline is more stable for the sparse and heterogeneous video feature space. The result does not necessarily imply that CatBoost is unsuitable for the task in general, but rather that the tested configuration is less robust to the particular mixture of sensor, text, audio, and visual features used in our pipeline.

%Overall, the video experiments show that performance is limited less by the absence of modalities and more by the \textbf{ difficulty of extracting reliable, task-aligned information from temporal content}. Adding speech transcripts, additional frames, or meme-oriented semantic indicators does not consistently improve performance; instead, the strongest configuration relies on the original feature set with top-300 feature selection, suggesting that compactness and noise control are more important than feature expansion for video-based sexism identification in this setting.
Overall, the video development experiments suggest that performance is limited less by the absence of modalities and more by the \textbf{difficulty of extracting reliable, task-aligned information from temporal content}. On the development split, adding speech transcripts, additional frames, or meme-oriented semantic indicators does not consistently improve performance; instead, the strongest configuration relies on the original feature set with top-300 feature selection. However, this conclusion is development-specific: as the official test results later show, the unfiltered video representation generalizes better on unseen data. This indicates that feature selection can be useful for suppressing validation noise, but may also remove weak signals that become useful under distribution shift.

\begin{table}[!ht]
\centering
\caption{Development results for Video Sexism Identification (Task 3.1). The table reports the ICM-Soft score for the optimal variant of each ablation step corresponding to the configurations defined in Table \ref{tab:video_experiments}.}
\label{tab:video_results}
\begin{tabular}{llc}
\hline
\textbf{ID} & \textbf{Configuration} & \textbf{ICM-Soft $\uparrow$ (Task 3.1)} \\
\hline
\textbf{E9} & BASE (Full feature space without selection) & -0.0032 \\
\textbf{E10}& \textbf{SELECT Top-$k$ features ($k=300$)} & \textbf{0.0378 (Best)} \\
\textbf{E11}& ADD Whisper transcription + 3-frame CLIP & -0.0200 \\
\textbf{E12}& REMOVE Whisper transcription & -0.0319 \\
\textbf{E13}& ADD LLM-derived semantic indicators & 0.0013 \\
\textbf{E14}& SUBSTITUTE CatBoost & -0.3588 \\
\hline
\end{tabular}
\end{table}

Following the same hierarchical strategy, utilizing the optimal E10 video configuration yields development scores of \textbf{-1.0200 for Task 3.2} and \textbf{-7.2493 for Task 3.3}.
\subsection{Submitted Runs}

Following the development-time evaluation, we structure our official submissions for the EXIST 2026 framework. We submit three distinct runs per subtask, covering both soft-label and hard-label regimes, resulting in 36 submissions in total. Our submission strategy is explicitly designed to contrast our baseline multimodal pipelines against our  feature-selected and LLM-enriched configurations.

\textbf{Task 2 (Meme Submissions):} To exploit the hierarchical normalization strategy, we combine different gatekeeper and downstream configurations by pairing different gatekeeper models with downstream classifiers based on the optimal experimental configurations identified in Table 3:

\begin{itemize}
    \item \textbf{Run 1 (Pragmatic-Driven):} Utilizes the compact pragmatic cues (E6) for both the binary (Task 2.1) and source intention (Task 2.2) subtasks, with Task 2.2 dynamically constrained by the Task 2.1 (E6) predictions. For sexism categorization (Task 2.3), it utilizes the comprehensive taxonomy-aligned predictions (E7),  clipped by the E6 gatekeeper.
    
    \item \textbf{Run 2 (Focused Semantic):} Utilizes the focused semantic cues (E8) for Tasks 2.1 and 2.2, with Task 2.2 constrained by the Task 2.1 (E8) gatekeeper. Categorization (Task 2.3) relies on E7 predictions, constrained by the E8 gatekeeper.
    
    \item \textbf{Run 3 (Contrastive Baseline \& Hybrid Gating):} For Task 2.1, this run utilizes the pure multimodal representation with feature selection (E3) without any LLM-derived features. For Task 2.2, it leverages E7 predictions constrained by the E8 gatekeeper (derived from Task 2.1). Finally, Task 2.3 uses the focused E8 configuration, again constrained by the E8 gatekeeper.
\end{itemize}

\textbf{Task 3 (Video Submissions):}
Given the video modality's sensitivity to cross-modal noise (as shown in Table \ref{tab:video_results}), our submissions  evaluate the effect of the dimensionality reduction strategy on the single-frame baseline (E9):
\begin{itemize}
    \item \textbf{Run 1 (Primary Best):} Trained exclusively on the optimally reduced space of the top 300 features (E10).
    \item \textbf{Run 2 (Moderate Reduction):} Trained on a broader, moderately reduced space of the top 400 features.
    \item \textbf{Run 3 (Unfiltered Baseline):} Trained on the complete initial feature space (E9), relying solely on algorithmic regularization without explicit feature selection.
\end{itemize}

The hard-label submissions are algorithmically derived directly from these soft-label probability distributions, strictly adhering to the thresholding rules and annotator-consensus logic detailed in Section \ref{sec:architecture}.

\subsection{Official Test Results}

Our official performance on the EXIST 2026 test set is summarized in Table \ref{tab:official_results}.The official evaluation platform evaluated all 36 of our submissions, covering three distinct runs per subtask across both the soft and hard evaluation regimes.

%Our proposed feature-engineered pipelines demonstrated highly competitive and stable performance, particularly in the soft evaluation setting and the more complex categorization tasks. 
%The official results show competitive performance on the downstream characterization subtasks, particularly in the soft evaluation setting, while binary meme sexism identification remains comparatively weaker. Specifically, our system achieves its best placements in the fine-grained sexism categorization for videos (Task 3.3), while also strong robustness is observed across the remaining downstream tasks (Tasks 2.2, 2.3, 3.1, and 3.2). 
The official results show stronger performance on the downstream characterization subtasks than on binary meme sexism identification, particularly in the soft evaluation setting. The best placements are obtained for fine-grained video sexism categorization (Task 3.3), while the remaining downstream tasks also show relatively stable rankings compared with Task 2.1. This pattern suggests that the proposed feature-based approach is more effective when the output space allows graded or multi-label characterization, whereas binary meme sexism identification remains more sensitive to ambiguous or borderline cases.
%For the binary meme sexism identification (Task 2.1), our variations achieved the 59th, 60th, and 61st positions.

As anticipated, the performance in the hard evaluation regime is comparatively lower across all runs. This discrepancy is a direct architectural consequence of our pipeline design: our models are optimized entirely on the continuous probability space (soft labels) using logistic regression objectives, with the discrete hard labels derived strictly post-hoc via thresholding and argmax operations.

\begin{table}[!t]
\centering \small
\caption{The table reports the official soft metrics (ICM-Soft, normalized ICM, cross-entropy where applicable) and hard metrics (ICM-Hard, normalized ICM, and F1).}
\label{tab:official_results}
\resizebox{\textwidth}{!}{
\begin{tabular}{lc|ccc|ccccc}
\hline
\multirow{2}{*}{\textbf{Subtask}} & \multirow{2}{*}{\textbf{Run}} & \multicolumn{4}{c|}{\textbf{Soft Evaluation}} & \multicolumn{4}{c}{\textbf{Hard Evaluation}} \\
 & & %\textbf{Rank} & 
 \textbf{ICM-Soft} & \textbf{Norm} & \textbf{CE} %& \textbf{Rank} 
 & \textbf{ICM-Hard} & \textbf{Norm} & \textbf{F1} \\
\hline
\multirow{3}{*}{\textbf{Task 2.1}} 
 & Run 1 & %60 & 
 -0.4675 & 0.4249 & 0.9282 & %75& 
 0.0354& 0.5180& 0.7011\\
 & Run 2 & %59 & 
 -0.4625 & 0.4257 & 0.9219 & %71& 
 0.0513& 0.5261& 0.7078\\
 & Run 3 & %61 & 
 -0.4753 & 0.4236 & 0.9223 & %67& 
 0.0605& 0.5308& 0.7089\\
\hline
\multirow{3}{*}{\textbf{Task 2.2}} 
 & Run 1 & %21& 
 -1.3450& 0.3570& 1.4486& %48& 
 -0.3446& 0.3802& 0.3913\\
 & Run 2 & %22& 
 -1.3525& 0.3562& 1.4343& %39& 
 -0.3082& 0.3929& 0.4033\\
 & Run 3 & %20& 
 -1.3266& 0.3689& 1.4405& %43& 
 -0.3272& 0.3863& 0.3945\\
\hline
\multirow{3}{*}{\textbf{Task 2.3}} 
 & Run 1 & %22& 
 -5.4961& 0.2087& -& %38& 
 -0.9530& 0.3023& 0.3501\\
 & Run 2 & %21& 
 -5.4766& 0.2098& -& %33& 
 -0.9207& 0.3090& 0.3502\\
 & Run 3 & %23& 
 -5.5438& 0.2068& -& %31& 
 -0.9125& 0.3107& 0.3482\\
\hline
\multirow{3}{*}{\textbf{Task 3.1}} 
 & Run 1 & %25& 
 -0.0201& 0.4965& 0.9314& %55& 
 0.0652& 0.5329& 0.6708\\
 & Run 2 & %24& 
 -0.0093& 0.4984& 0.9429& %51& 
 0.0850& 0.5429& 0.6645\\
 & Run 3 & %21& 
 0.0349& 0.5061& 0.9374& %53& 
 0.0788& 0.5397& 0.6589\\
\hline
\multirow{3}{*}{\textbf{Task 3.2}} 
 & Run 1 & %21& 
 -1.7564& 0.3129& 1.2992& %38& 
 -0.0588& 0.4778& 0.5751\\
 & Run 2 & %25& 
 -1.8038& 0.3079& 1.3146& %40& 
 -0.0696& 0.4737& 0.5752\\
 & Run 3 & %20& 
 -1.5763& 0.3321& 1.2847& %36& 
 -0.0276& 0.4896& 0.5781\\
\hline
\multirow{3}{*}{\textbf{Task 3.3}} 
 & Run 1 & %16& 
 -6.5768& 0.1077& -& %23& 
 -0.3731& 0.3793& 0.2978\\
 & Run 2 & %17& 
 -6.5801& 0.1075& -& %27& 
 -0.3877& 0.3746& 0.2902\\
 & Run 3 & %15& 
 -6.5497& 0.1094& -& %25& 
 -0.3812& 0.3766& 0.2866\\
\hline
\end{tabular}
}
\end{table}
Overall, the performance on the official test set provides critical insights into the generalization capabilities of our proposed methods, revealing distinct behaviors between the static and dynamic modalities.

%For the meme modality (Task 2), the test rankings support the usefulness of our LLM-enriched hierarchical strategy. In the binary identification subtask (Task 2.1), the focused semantic configuration (E8, submitted as Run 2) outperforms both the pragmatic variant (E6, Run 1) and the contrastive multimodal baseline lacking LLM features (E3, Run 3). %This confirms that high-level semantic indicators successfully aid the model in generalizing to unseen data. 
For the meme modality (Task 2), the official results support the usefulness of LLM-derived semantic indicators, but also show that their utility depends on the subtask. In binary sexism identification (Task 2.1), the focused semantic configuration (E8, submitted as Run 2) obtains the best result among our submitted runs, outperforming both the pragmatic variant (E6, Run 1) and the contrastive multimodal baseline without LLM-derived features (E3, Run 3). This suggests that focused high-level indicators such as stereotyping, objectification, and misogyny improve generalization for binary meme sexism identification. For source-intention detection (Task 2.2), however, Run 3 achieves the best soft ranking among our submissions. This suggests that stance prediction may benefit from broader downstream semantic information rather than only focused binary sexism cues. This is consistent with the nature of the task: distinguishing \textbf{Direct} from \textbf{Judgemental} requires modeling communicative function, authorial stance, and whether sexist content is endorsed, quoted, reported, or criticized. For sexism categorization (Task 2.3), Run 2 achieves the best soft result among our submissions. This run combines comprehensive taxonomy-aligned predictions (E7) with the focused E8 gatekeeper, suggesting that different levels of semantic granularity are useful at different points of the hierarchy. Focused cues are effective for estimating whether sexism is present, while broader taxonomy-aligned cues appear more suitable for allocating probability mass across fine-grained, partially overlapping sexism categories.
%This suggests that high-level semantic indicators improve generalization for binary meme sexism identification. Furthermore, the multi-label categorization results (Task 2.3) underscore the importance of aligning feature granularity with task complexity. Our highest ranking for this subtask is achieved by Run 2, which utilizes the comprehensive taxonomy-aligned predictions (E7) constrained by the focused E8 gatekeeper. Conversely, utilizing the narrower E8 configuration for both gating and fine-grained categorization (Run 3) yields a lower rank, indicating that while focused semantic cues excel at binary detection, %a broader structural taxonomy is strictly required to effectively disentangle multi-label sexism categories.
%a broader taxonomy-aligned representation appears more suitable for disentangling multi-label sexism categories.

For the video modality (Task 3), the official rankings reveal a clear dev--test discrepancy regarding feature selection. During development, the top-300 selected feature subset achieves the best Task 3.1 score, while the unfiltered baseline underperforms. On the official test set, however, this trend reverses: across all three video subtasks, the unfiltered baseline (Run 3) obtains the best ranking among our submitted video runs. 
%This suggests that aggressive feature selection may have overfit to the development distribution in the sparse and heterogeneous video setting. 
This suggests that aggressive feature selection may have specialized too strongly to the development distribution in the sparse and heterogeneous video setting.
A plausible explanation is that the complete representation preserves weak but useful signals that were removed during feature selection. In short-form videos, relevant evidence may be distributed across visual frames, OCR text, acoustic descriptors, and sensor-derived features, and some of these signals may be individually weak but collectively useful. The official results therefore indicate that video modeling requires not only noise control, but also robustness to distribution shift and better temporal alignment of the extracted modalities.
%For the video modality (Task 3), the test rankings expose a significant generalization gap regarding our dimensionality reduction strategy. During the development phase, aggressive feature selection appears strictly optimal, while the unfiltered baseline (E9) heavily underperforms. However, on the official test set, this dynamic  reverses: across all three video subtasks, the unfiltered baseline (Run 3) consistently achieves the highest global rankings (21st, 20th, and 15th, respectively), clearly outperforming the feature-selected variants (Runs 1 and 2). %This indicates that aggressive feature selection in the highly sparse and cross-modal video domain may have overfit to the development distribution on the development distribution. 
%This suggests that aggressive feature selection in the sparse and heterogeneous video domain may have overfit to the development distribution.
%The complete, unselected representation inherently preserves critical variance, allowing the algorithmic regularization of the gradient boosting architecture to generalize much more robustly to the unseen test distribution.
%A plausible explanation is that the complete, unselected representation preserves weak but useful signals that were removed by feature selection, allowing the regularized gradient boosting model to generalize better to the unseen test distribution.

All in all, the development and official results indicate that the optimal feature granularity depends on both modality and subtask. For memes, compact semantic indicators are most useful for binary sexism identification, whereas broader taxonomy-aligned indicators are more useful for fine-grained categorization. For videos, development-time feature selection improves validation performance but does not generalize as well as the unfiltered representation on the official test set. This suggests that static meme understanding benefits from semantic abstraction, while short-form video understanding is limited by the robustness and temporal alignment of the extracted signals.
 
\section{Analysis and Discussion}

The development and official results suggest two main observations. First, meme-based sexism identification benefits from\textbf{ semantic abstraction}: direct visual embeddings and compact LLM-derived indicators help expose implicit cues such as stereotyping, objectification, and misogyny that are not always recoverable from OCR text alone. Second, video-based sexism identification is more sensitive to the \textbf{reliability and temporal alignment }of the extracted signals. Adding more information sources does not necessarily improve performance when frames, speech, OCR, audio, and sensor-derived features are weakly aligned with the moment where sexist meaning is expressed. We discuss these findings through modality contribution, error patterns, feature importance, source-intention ambiguity, category overlap, and uncertainty.

\subsection{Modality Contribution and Generalization}

The ablation results show that the contribution of each modality depends strongly on the structure of the input. For memes, visual information provides important complementary evidence. In explicit cases, OCR text may be sufficient, especially when the meme contains direct sexist or misogynistic language. However, many memes rely on the interaction between the image and the overlaid text: the image may establish the target, social role, stereotype, or humorous setup required to interpret the caption. This helps explain the improvement from the initial BLIP--OCR baseline (E1) to the CLIP-enriched representation (E2). CLIP embeddings appear to preserve visual cues that are not fully captured by generated image descriptions or OCR text alone.

For videos, the development experiments initially suggest that feature selection is beneficial: the top-300 feature subset improves over the unselected video baseline. However, the official test results show that this conclusion does not fully transfer to unseen data, since the unfiltered video run generalizes better across all three video subtasks. This dev--test discrepancy suggests that feature selection may have captured validation-specific regularities rather than robust video-level cues. A plausible explanation is that the complete feature space preserves weak but useful signals that were removed by aggressive selection.

This contrast indicates that video performance is not limited simply by the absence of modalities, but by the difficulty of extracting temporally aligned and task-relevant information from short-form videos. Whisper transcripts, multiple sampled frames, OCR text, acoustic descriptors, and sensor-derived features may each contain useful information, but they may correspond to different temporal moments. Without explicit temporal modeling, additional modalities can introduce noise or dilute the relevant signal. Thus, the main limitation of the video pipeline is not multimodality itself, but the lack of alignment between modalities and the specific moment where sexist meaning is expressed.

\subsection{Offensiveness vs. Sexism}

A prevalent failure mode in the soft-label predictions is the model's difficulty in disentangling \textbf{generic offensiveness from targeted sexism}. Qualitative error analysis reveals two distinct patterns. First, the model frequently produces false positives when processing highly offensive language lacking a gendered target. For instance, a meme containing the text \textit{``YOU SIR... ARE A BITCH''} yields a predicted sexism probability of 0.863. This prediction suggests that the model relies heavily on lexical offensiveness and fails to sufficiently condition on the male-directed target signaled by \textit{``Sir''}. In such cases, the presence of derogatory or sexualized vocabulary is treated as evidence of sexism, even though offensiveness alone does not necessarily imply gender-directed discrimination.

Second, the model exhibits false negatives in cases where sexist stereotypes are conveyed through polite, humorous, or metaphorical language without explicit profanity. For example, a meme stating \textit{``I CANNOT COMMENT ON YOUR STEPMOM COWS ARE SACRED...''} receives a predicted probability of only 0.126. Because the text lacks aggressive vocabulary, the base textual and multimodal embeddings appear insufficient to capture the implicit objectification and derogatory comparison. This type of error highlights why higher-level semantic indicators are useful: they can make latent cues such as objectification, stereotyping, or misogynistic framing more explicit to the downstream model.

These two patterns show that sexism detection cannot be reduced to toxicity detection. A system that over-relies on offensive language risks over-predicting sexism in generally toxic content, while a system that underweights implicit stereotypes risks missing socially normalized or humorous sexist expressions. The strongest configurations are therefore those that combine low-level multimodal evidence with semantic cues that encode the target, stance, and social meaning of the content.

\subsection{Interpretability and Feature Importance}

To better understand the behavior of the gradient boosting models, we inspected XGBoost feature importance scores across the meme configurations. The analysis indicates that the LLM-derived semantic indicators are not merely auxiliary features. In the strongest meme settings, indicators related to stereotyping, objectification, ideological inequality, and misogyny rank among the most influential dimensions. This does not constitute a causal explanation of individual predictions, since tree-based feature importance can be affected by feature correlations and split structure. However, it shows that the fitted models consistently make use of the semantic indicators when they are available.

This helps explain why compact semantic augmentation improves meme performance. Dense visual and textual embeddings encode broad information, but they are not explicitly aligned with the EXIST taxonomy. By contrast, the LLM-derived indicators act as \textbf{weak semantic annotators}: they compress the visual-textual content into interpretable, label-aligned concepts. This representation is particularly suitable for gradient boosting models, which can exploit sparse semantic cues more directly than high-dimensional embedding spaces.

The results also show that not all semantic indicators are equally useful. Broad or weakly targeted cues, such as generic irony or male-directed insults, were less informative in development experiments. Focused cues related to stereotyping, objectification, and misogyny were more useful for binary meme sexism identification, while the broader taxonomy-aligned feature set was more useful for fine-grained categorization. This suggests that semantic prompting should be task-specific: compact indicators are preferable when the goal is to detect whether sexism is present, whereas broader category-aligned indicators are more useful when the goal is to distribute probability mass across overlapping sexism categories.

\subsection{Source-Intention Ambiguity}

The source-intention subtasks are difficult because they require more than detecting sexist content; they require identifying the stance of the source toward that content. The distinction between \textbf{Direct} and \textbf{Judgemental} often depends on whether sexist language is endorsed, quoted, condemned, reported, or used ironically. A model may correctly detect the presence of sexist vocabulary or imagery, but still fail to determine whether the content promotes sexism or criticizes it.

This explains why simple irony-related features were insufficient. Irony is only one mechanism by which stance can be reversed. Judgemental content may contain explicit sexist expressions precisely because it is describing or condemning them, while direct sexist content may express discriminatory beliefs without using overtly aggressive language. Therefore, source intention is better understood as a \textbf{stance-attribution problem} rather than a simple extension of binary sexism detection.

The hierarchical design partially addresses this issue by using the binary sexism score as a gate for downstream intention prediction. However, the remaining difficulty of Tasks 2.2 and 3.2 suggests that future systems should model stance more explicitly. Useful directions include features that distinguish use from mention, detect quotation or reporting structures, identify condemnation markers, and capture whether the authorial voice aligns with or distances itself from the sexist content.

\subsection{Category-Level Error Analysis}

The fine-grained categorization subtasks are affected by strong dependencies among the official sexism categories. The co-occurrence matrix shows that categories such as \textbf{Stereotyping and Dominance}, \textbf{Ideological Inequality}, and \textbf{Objectification} frequently overlap. This \textbf{overlap should not be interpreted only as annotation noise}. It reflects the structure of sexism as a social phenomenon: stereotypes often support ideological inequality, objectification may co-occur with sexualized forms of violence, and misogynistic expressions may involve dominance, harassment, or dehumanization.

This helps explain why fine-grained categorization is harder than binary identification. The model is not simply choosing among independent labels; it must assign scores to partially overlapping categories. A meme can simultaneously express a stereotype, reinforce unequal gender roles, and objectify women. Under such conditions, hard category boundaries are inherently unstable, and multi-label soft prediction is more appropriate than forcing mutually exclusive decisions.

The stronger performance of taxonomy-aligned semantic features for Task 2.3 is consistent with this interpretation. The broader semantic feature set provides category-specific evidence while still allowing overlap through multi-label prediction. Rather than fully separating highly correlated concepts, these features give the model explicit signals for each category, helping it allocate probability mass more meaningfully across the taxonomy.

\subsection{Disagreement, Confidence, and Threshold Instability}

The confidence analysis provides additional evidence that sexism identification contains many uncertain boundary cases. Using the out-of-fold predictions of the strongest meme model, correct predictions show higher mean confidence than incorrect predictions. True positives and true negatives have mean confidence values of 70.6\% and 66.1\%, respectively, while false positives and false negatives drop to 64.8\% and 61.1\%. This suggests that the model is less confident on harder or more ambiguous examples. However, this should not be interpreted as full calibration; a complete calibration analysis would require reliability curves or expected calibration error.

The boundary-case analysis is particularly important. Instances with predicted probabilities between 0.4 and 0.6 account for 1,278 examples, or approximately 32.1\% of the dataset. \textbf{Hard-label accuracy on these cases drops to 52.4\%}, compared with 64.6\% overall. This shows that a large portion of the data lies near the decision boundary, where small probability changes can flip the hard prediction.

This result supports the Learning with Disagreement setup of EXIST. In socially sensitive tasks, uncertainty is not merely model failure or dataset noise; it often reflects genuine ambiguity in interpretation. A meme may be offensive but not clearly sexist, sexist but expressed indirectly, or judgemental rather than direct. Soft-label evaluation is therefore better aligned with the nature of the task, because it allows systems to express uncertainty instead of forcing all examples into discrete decisions. The lower performance of hard submissions is consistent with this observation: \textbf{post-hoc thresholding collapses meaningful uncertainty} into brittle binary or categorical choices.

\section{Limitations}

Several limitations follow from the analysis above. First, the LLM-derived semantic indicators are prompt-dependent. They provide useful high-level abstractions for memes, but they also inherit the biases, blind spots, and prompt sensitivity of the specific underlying language model utilized for feature extraction (i.e., Qwen2.5-1.5B-Instruct). If the LLM misinterprets a meme, the resulting semantic feature may propagate this error to the downstream classifier.

Second, the feature-importance analysis should be interpreted cautiously. High importance scores indicate that the XGBoost models heavily rely on the semantic indicators, but they do not prove that these features are causally responsible for each prediction. Feature importance can be affected by correlations among features and by the inherent structure of the fitted decision trees.

Third, the video pipeline lacks explicit temporal modeling. The system represents videos through aggregated or sampled features, such as representative frames, OCR text, acoustic descriptors, and sensor-derived metadata. This makes the pipeline efficient, but it may discard the temporal sequence in which sexist meaning emerges. The dev--test reversal for video feature selection further suggests that validation-based feature reduction may be unstable when video signals are sparse, heterogeneous, or weakly aligned.

Finally, the hard-label conversion strategy remains brittle. Fixed thresholds such as 0.5 for binary sexism identification and 0.33 for category selection do not fully reflect the ambiguity observed in boundary cases. Future work should therefore explore calibration-aware thresholding, uncertainty-aware decision rules, and stronger use of annotator-disagreement information. A more detailed analysis of physiological and sensor-derived signals may also help determine whether user responses are especially informative for ambiguous or high-disagreement examples.

\section{Conclusion}

In this paper, we present the AILS-NTUA system for EXIST 2026, addressing sexism identification, source-intention detection, and sexism categorization in memes and TikTok videos. Our approach combines pretrained multimodal representations, demographic and sensor-derived features, LLM-derived semantic indicators, and gradient-boosted regression models within a hierarchical late-fusion framework.
Extensive development experiments show that CLIP image embeddings and focused LLM-derived semantic cues improve meme sexism identification, while taxonomy-aligned cues are more valuable for fine-grained categorization. For videos, development results initially favor aggressive feature selection, but official test results demonstrate that the unfiltered representation generalizes better, revealing the risk of overfitting in sparse and noisy multimodal video settings.
Overall, our findings suggest that targeted semantic feature engineering can be effective for static multimodal sexism detection, whereas short-form video content requires more robust temporal and cross-modal modeling. Future work will focus on stronger video representations, calibration-aware thresholding, and a more detailed analysis of how physiological signals relate to annotator disagreement and model uncertainty.

%\section{Acknowledgments}

\bibliography{sample-ceur}

\appendix
\section{LLM Prompting Templates}
\label{appendix:prompts}

In this appendix, we provide the exact prompts used to query the Large Language Models for extracting structural, semantic, and taxonomy-aligned features from the dataset. The variable \texttt{\{super\_sentence\}} is a concatenation of the image description (generated via BLIP) and the explicitly transcribed on-screen text (via OCR).

\subsection{Base Semantic Features Prompt}
This prompt was designed to extract three foundational socio-linguistic cues: the presence of irony, targeted insults towards men, and baseline stereotypes.

\begin{verbatim}
Analyze the following meme. You are given a description of the
image and the text written on it (which may be in English, Spanish, 
or other languages). You must answer exactly with three digits 
separated by commas (e.g., 1,0,1).

{super_sentence}

Questions:
1. Considering both the image and the text, is there irony, 
   sarcasm, or a joke? (1 for Yes, 0 for No)
2. Is there an insult, curse word, or derogatory language directed 
   specifically at a man/boy? (1 for Yes, 0 for No)
3. Does the meme contain traditional stereotypes about women (e.g., 
   kitchen, driving, emotional)? (1 for Yes, 0 for No)

Output ONLY the three digits separated by commas. No other words.
\end{verbatim}

\subsection{Comprehensive 10-Feature Taxonomy Prompt}
This prompt was utilized in the advanced experiments (e.g., E7) to force the LLM to evaluate the input across all fine-grained taxonomy dimensions simultaneously, helping to decouple highly entangled semantic categories.

\begin{verbatim}
Analyze the following meme. You are given a description of the
image and the text written on it. You must answer exactly with ten 
digits separated by commas (e.g., 1,0,0,1,0,1,0,0,1,0).

{super_sentence}

Questions:
1. IRONY: Is there irony, sarcasm, or a joke? (1 for Yes, 0 for No)
2. MALE INSULT: Is there an insult or derogatory language directed 
   specifically at a man/boy? (1 for Yes, 0 for No)
3. CONTRADICTION: Is there a contrast or contradiction between the 
   neutral text and the visual that creates a sexist joke? (1/Yes, 0/No)
4. DIRECT: Is the intention to write a message that is sexist by itself 
   or incites to be sexist? (1 for Yes, 0 for No)
5. JUDGEMENTAL: Is the intention to judge, describing sexist situations 
   or behaviours with the aim of condemning them? (1 for Yes, 0 for No)
6. IDEOLOGICAL AND INEQUALITY: Does the meme discredit the feminist
   movement, reject inequality between men and women, or present men as
   victims of gender-based oppression? (1 for Yes, 0 for No)
7. STEREOTYPING AND DOMINANCE: Does the meme express false ideas
   suggesting women are more suitable for certain roles (mother, 
   submissive, etc.), inappropriate for certain tasks (driving, etc.), 
   or claim men are superior? (1 for Yes, 0 for No)
8. OBJECTIFICATION: Does the meme present women as objects, focus on
   compliance with beauty standards, hypersexualize female attributes, 
   or put women's bodies at the disposal of men? (1 for Yes, 0 for No)
9. SEXUAL VIOLENCE: Are there sexual suggestions, requests for sexual 
   favors, or harassment of a sexual nature? (1 for Yes, 0 for No)
10. MISOGYNY AND NON-SEXUAL VIOLENCE: Does the meme express hatred
    and non-sexual violence towards women? (1 for Yes, 0 for No)

Output ONLY the ten digits separated by commas. No other words.
\end{verbatim}

\subsection{Targeted Fine-Grained Prompt ("The Holy Trinity")}
This streamlined prompt focuses exclusively on the most critical and frequently overlapping fine-grained categories, utilized in our highest-performing pipelines to maximize classification confidence.

\begin{verbatim}
Analyze the following meme. You are given a description of the image 
and the text written on it. You must answer exactly with three digits 
separated by commas (e.g., 1,0,1).

{super_sentence}

Questions:
1. STEREOTYPING AND DOMINANCE: Does the meme express false ideas 
   suggesting women are more suitable for certain roles (mother, 
   submissive, etc.), inappropriate for certain tasks (driving, etc.), 
   or claim men are superior? (1 for Yes, 0 for No)
2. OBJECTIFICATION: Does the meme present women as objects, focus on 
   compliance with beauty standards, hypersexualize female attributes, 
   or put women's bodies at the disposal of men? (1 for Yes, 0 for No)
3. MISOGYNY AND NON-SEXUAL VIOLENCE: Does the meme express hatred and 
   non-sexual violence towards women? (1 for Yes, 0 for No)

Output ONLY the three digits separated by commas. No other words.
\end{verbatim}

\section{Use of AI tools declaration}
AI tools were used for manuscript polishing, clarity enhancement and typo/grammar/syntax fixing.

\end{document}